\documentclass[letterpaper, 10 pt, conference]{ieeeconf}  

\IEEEoverridecommandlockouts                              

\usepackage{amsmath, amssymb}
\usepackage{graphicx}
\usepackage{algorithm}
\usepackage{algpseudocode}
\usepackage{cite}

\title{\LARGE \bf
A Robust and Energy-Efficient Trajectory Planning Framework for High-Degree-of-Freedom Robots
}

\author{
    Sajjad Hussain$^{1}$\thanks{*Corresponding author. Email: s.hussain4@brighton.ac.uk}, 
    Md Saad$^{2}$\thanks{*Corresponding author. Email: mohammed.saadsalik@gmail.com}, 
    Almas Baimagambetov$^{1}$, 
    Khizer Saeed$^{1}$\\
    \vspace{0.2cm}
    \\
    $^{1}$School of Architecture, Technology and Engineering, University of Brighton, Brighton, UK\\
    $^{2}$Department of Mechanical Engineering, Jamia Millia Islamia, New Delhi, India.\\
}
\begin{document}

\maketitle
\thispagestyle{empty}
\pagestyle{empty}

\begin{abstract}
Energy efficiency and motion smoothness are essential in trajectory planning for high-degree-of-freedom robots to ensure optimal performance and reduce mechanical wear. This paper presents a novel framework integrating sinusoidal trajectory generation with velocity scaling to minimize energy consumption while maintaining motion accuracy and smoothness. The framework is evaluated using a physics-based simulation environment with metrics such as energy consumption, motion smoothness, and trajectory accuracy. Results indicate significant energy savings and smooth transitions, demonstrating the framework's effectiveness for precision-based applications. Future work includes real-time trajectory adjustments and enhanced energy models.
\end{abstract}

\section{INTRODUCTION}
High-degree-of-freedom robots are widely used in complex tasks, such as assembly lines, material handling, and surgical operations. However, these tasks are energy-intensive and create challenges for long-term deployments \cite{martyushev2023review}. The traditional trajectory planning methods usually focus on accuracy and speed but do not consider energy efficiency, which leads to excessive energy consumption and mechanical stress. This paper addresses the challenges by proposing a trajectory planning framework that integrates non-linear sinusoidal trajectories with velocity scaling to minimize energy usage while preserving motion smoothness and accuracy.

\section{Related Work}
Recent research in energy-efficient trajectory planning for high-degree-of-freedom (DOF) robots has focused on reducing energy consumption while maintaining performance. Wang et al. (2024) proposed a deep reinforcement learning-based framework for real-time trajectory optimization, achieving significant energy savings in industrial robots by dynamically learning optimal paths \cite{wang2024energy}. Zhang et al. (2022) emphasized integrating kinematic and dynamic constraints for optimal energy-efficient trajectories, transforming the planning problem into a nonlinear optimization task \cite{li2023energy}. Furthermore, recent advancements in trajectory generation leverage machine learning techniques to handle high-dimensional data, balancing efficiency and smoothness in robotic systems. These methods collectively address energy optimization challenges, offering practical solutions for real-world robotic applications.

\section{Proposed Framework}
\subsection{System Overview}
\begin{figure}[ht]
    \centering
    \includegraphics[width=0.9\linewidth]{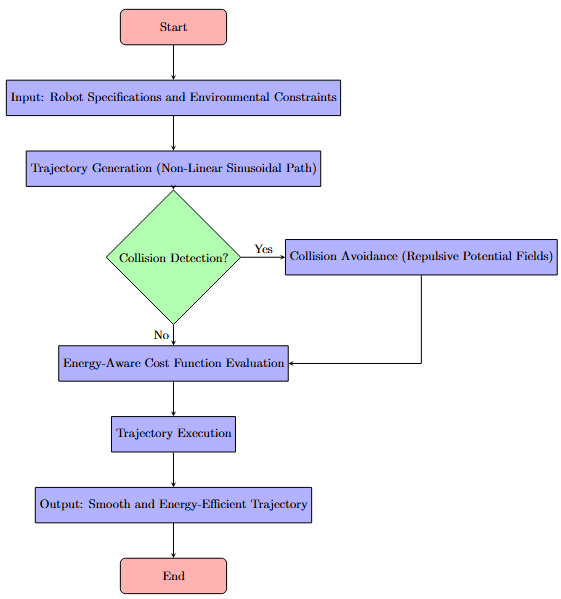}
    \caption{Framework for Energy-Efficient Trajectory Planning.}
    \label{fig:framework_diagram}
\end{figure}

The proposed framework consists of three key components: Energy-Aware Cost Function, Adaptive Trajectory Generation, and Collision Avoidance. The high-level flow of the framework is depicted in Figure~\ref{fig:framework_diagram}, which demonstrates the interaction between the different modules and their contribution to the overall trajectory planning process.

\subsection{Energy-Aware Cost Function}
The cost function evaluates energy consumption as:
\begin{equation}
    C_{\text{energy}} = \int_0^T \left( \sum_{i=1}^{n} \left( \tau_i^2 + \lambda \dot{q}_i^2 \right) \right) dt,
\end{equation}
where $\tau_i$ represents joint torques, $\dot{q}_i$ is joint velocity, and $\lambda$ is a weighting factor.

\subsection{Adaptive Trajectory Generation}
Trajectories are generated using cubic spline interpolation:
\begin{equation}
    q_i(t) = a_i + b_i t + c_i t^2 + d_i t^3, \quad t \in [t_0, t_f].
\end{equation}
This ensures smooth transitions while adhering to robot constraints.

\subsection{Collision Avoidance}
Real-time collision avoidance modifies trajectories using a repulsive potential field:
\begin{equation}
    F_{\text{repulsive}} = 
    \begin{cases} 
    k \left( \frac{1}{d} - \frac{1}{d_{\text{safe}}} \right) \frac{1}{d^2} \hat{n}, & \text{if } d \leq d_{\text{safe}}, \\
    0, & \text{if } d > d_{\text{safe}}.
    \end{cases}
\end{equation}

\section{Implementation}
\subsection{Simulation Setup}
The framework was implemented and validated using a physics-based simulation platform, PyBullet. A 7-degree-of-freedom robotic manipulator was modeled with realistic kinematic and dynamic constraints. The simulation incorporated environmental factors such as dynamic obstacles, workspace boundaries, and collision constraints. Robot specifications, environmental parameters, and operational constraints were integrated to validate trajectory planning and execution. 

\subsection{Evaluation Metrics}
The proposed framework is evaluated using key performance metrics to assess its effectiveness in optimizing trajectory planning for high-degree-of-freedom robots. Table I summarizes the observations for each metric, highlighting energy efficiency, motion smoothness, and trajectory accuracy.

\begin{table}[h!]
\centering
\caption{Summary of Evaluation Metrics and Observations}
\label{tab:evaluation_metrics}
\begin{tabular}{|p{3.5cm}|p{4.5cm}|}
\hline
\textbf{Metric} & \textbf{Observation} \\
\hline
\textbf{Smooth Energy Consumption} & Energy consumption decreases progressively over time, demonstrating effective velocity scaling and optimized trajectory planning. \\
\hline
\textbf{Acceleration Over Time} & Periodic peaks and valleys align with sinusoidal trajectory changes, ensuring smooth transitions and minimal abrupt motion variations. \\
\hline
\textbf{Cumulative Energy Consumption} & Displays a steady and consistent increase, indicating efficient energy usage without abrupt spikes or inconsistencies. \\
\hline
\textbf{Velocity Magnitude Over Time} & Periodic oscillations in velocity magnitude align with the sinusoidal trajectory, confirming smooth and controlled robotic motion. \\
\hline
\end{tabular}
\end{table}

\section{Results and Discussion}

\begin{figure}[h!]
    \centering
    \includegraphics[width=0.9\linewidth]{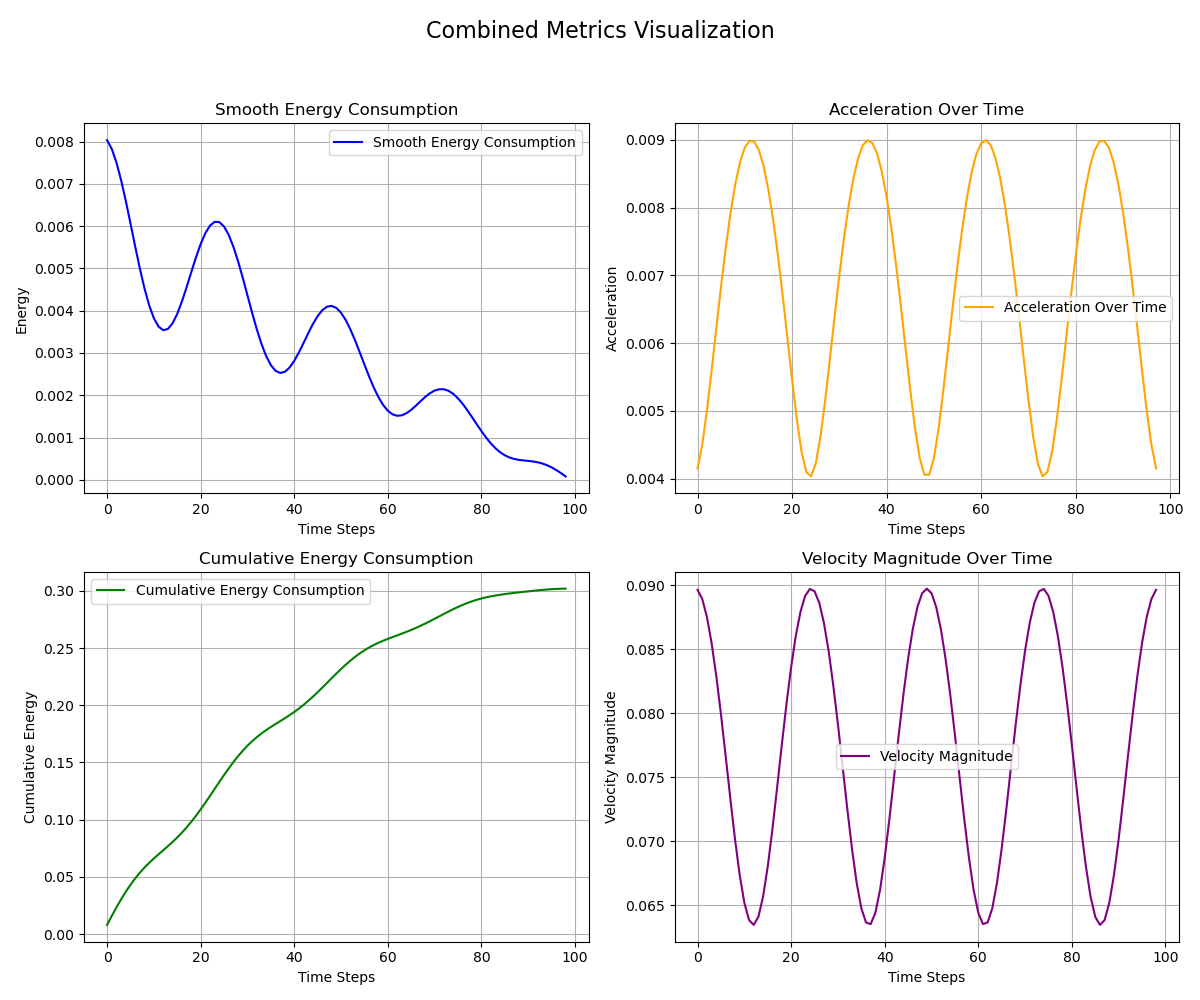}
    \caption{Combined Metrics Visualization: Smooth Energy Consumption, Acceleration, Cumulative Energy, and Velocity Magnitude.}
    \label{fig:metrics_visualization}
\end{figure}

The proposed trajectory planning framework demonstrates significant improvements in energy efficiency and motion smoothness. Key observations from the evaluation are summarized below:

\begin{itemize}
    \item \textbf{Smooth Energy Consumption:} The energy consumption decreases progressively over time, as shown in the top-left subplot of Figure~\ref{fig:metrics_visualization}, validating the effectiveness of velocity scaling.
    \item \textbf{Acceleration Over Time:} The acceleration graph (top-right) highlights periodic peaks and valleys, ensuring smooth transitions and minimal abrupt motion variations.
    \item \textbf{Cumulative Energy Consumption:} The bottom-left subplot reflects a steady and consistent increase in energy usage without abrupt spikes, confirming efficient trajectory planning.
    \item \textbf{Velocity Magnitude Over Time:} The bottom-right graph shows periodic oscillations in velocity magnitude, aligning with the sinusoidal trajectory for smooth and controlled motion.
\end{itemize}

The results underline the robustness of the proposed framework in achieving energy-efficient and precise motion for high-degree-of-freedom robotic systems. Future work will focus on real-time implementation and validation in real-world environments.

\section{Conclusion}
The proposed framework achieves energy-efficient and smooth trajectory planning for high-DOF robots using sinusoidal paths and velocity scaling. Results show improved energy efficiency and motion accuracy, highlighting its potential for industrial and precision robotics applications.

\end{document}